\documentclass{article}
\usepackage{ijcai18}

\usepackage{times}
\usepackage{stfloats}
\usepackage{xcolor}
\usepackage{soul}
\usepackage[utf8]{inputenc}
\usepackage[small]{caption}
\usepackage{times}
\usepackage{helvet}
\usepackage{courier}
\usepackage{url}  %Required
\usepackage{graphicx}  %Required
\usepackage{mwe}
\usepackage{subfig}
\usepackage{todonotes}
\usepackage{amsmath, bm, amssymb}
\DeclareMathOperator{\Expect}{\mathbb{E}}
\graphicspath{{../}{figures/}}

\usepackage[ruled,vlined]{algorithm2e}\usepackage{algorithmic}

\newcommand{\defterm}{\textbf}

\newcommand{\TTuple}[1][0.0ex]{\vec{t}\hspace{#1}}

\newcommand{\UTuple}[1][0.0ex]{\vec{u}\hspace{#1}}

\newcommand{\VTuple}{\vec{v}}
\newcommand{\Mrange}[1]{\ifthenelse{\equal{#1}{T}}{\TTuple_m}{\ifthenelse{\equal{#1}{U}}{\UTuple_m}{\ifthenelse{\equal{#1}{V}}{\VTuple_m}{\mbox{UNKNOWN
TERM ID}}}}}
\newcommand{\Prange}[1]{\ifthenelse{\equal{#1}{T}}{\vec{t}_{pa}}{\ifthenelse{\equal{#1}{U}}{\vec{u}_{pa}}{\ifthenelse{\equal{#1}{V}}{\vec{v}_{pa}}{\mbox{UNKNOWN
TERM ID}}}}}

\newcommand{\GroundPrange}[1]{\ifthenelse{\equal{#1}{T}}{\vec{t}_{pa,\grounding'}}{\ifthenelse{\equal{#1}{U}}{\vec{u}_{pa,\grounding'}}{\ifthenelse{\equal{#1}{V}}{\vec{v}_{pa,\grounding'}}{\mbox{UNKNOWN
TERM ID}}}}}

\newcommand{\grounding}{\gamma}

\newcommand{\feature}{x} % feature or desc attribute of object or link
\newcommand{\features}{\set{\feature}} % features 
\def\set#1{\mathbf{#1}}

\newcommand{\action}{a}
\newcommand{\home}{\it{Home}}
\newcommand{\away}{\it{Away}}
\newcommand{\none}{\it{Neither}}
\newcommand{\team}{\it{team}}
\newcommand{\egoal}{\it{goal}}
\newcommand{\sequences}{D}

\newcommand{\goal}{g}
\newcommand{\history}{s}

\newcommand{\pcounts}[4]{n^{#1}_{#4}(#2,#3)}
\newcommand{\Qt}[3]{Q^{#1}(#2,#3)}

\newcommand{\impact}[2]{\it{impact}^{#1}(#2)}
\newcommand{\Qmodel}[3]{\hat{Q}_{#1}(#2,#3)}
\newcommand{\tracel}{\it{tl}}

\title{IJCAI--18 Formatting Instructions\thanks{These match the formatting instructions of IJCAI-07. The support of IJCAI, Inc. is acknowledged.}}

\author{Jêröme Lang\\ 
Laboratoire d'Analyse et Modélisation des Systèmes pour l'Aide à la Décision (LAMSADE)  \\
pcchair@ijcai-18.org}

\begin{document}
\title{Deep Reinforcement Learning in Ice Hockey \\ for Context-Aware %Action Values and 
Player Evaluation
% Context-Aware Evaluations of Actions and Players in Ice Hockey Using Deep Reinforcement Learning
}

\author{
Guiliang Liu \and
Oliver Schulte \\
Simon Fraser University, Burnaby, Canada\\
gla68@sfu.ca,
oschulte@cs.sfu.ca
}

\maketitle
\begin{abstract}
% Emphasize our novelty, first work that apply Deep RL in sports games. 
A variety of machine learning models have been proposed to assess the performance of players in professional sports. However, they  have only a limited ability to model how player performance depends on the game context. This paper proposes a new approach to capturing  game context: we apply Deep Reinforcement Learning (DRL) to learn an action-value Q function from 3M play-by-play events in the National Hockey League (NHL). The neural network representation 
%of our DRL model 
integrates both continuous context signals and game history, using a possession-based LSTM. The learned Q-function is used to value players' actions under different game contexts. To assess a player's overall performance, we introduce a novel  Game Impact Metric (GIM) that aggregates the values of the player's actions. Empirical Evaluation shows GIM is consistent throughout a play season, and correlates highly with standard success measures  and future salary. 
\end{abstract}

%\listoftodos
\section{Introduction: Valuing Actions and Players}

With the advancement of high frequency optical tracking and object detection systems, more and larger event stream datasets for sports matches have become available. There is increasing opportunity for large-scale machine learning to model complex sports dynamics. Player evaluation is a major task for sports modeling that draws attention from both fans and team managers, who want to know which players to draft, sign or trade. Many models have been proposed~\cite{Buttrey2011,Macdonald2011,decroos2018actions,kaplan}.
%jones2011responses,
%\todo{cite van Haaren's latest?} \textcolor{blue}{do we know the name of that paper, can't google it} 
The most common approach has been to quantify the value of a player's action, and to evaluate players by the total value of the actions they took~\cite{Schuckers2013,mchale2012development}. 

However, traditional sports models assess only the actions that have immediate impact on goals (e.g. shots), but not the actions that lead up to them (e.g. pass, reception). And action values are assigned taking into account only a limited context of the action. But in realistic professional sports, the relevant context  is very complex, including game time, position of players, score and manpower differential, etc. 
%But only very limited of them have been molded. 

Recently, Markov models have been used to address these limitations. \cite{Routley2015a} used states of a Markov Game Model to capture game context and compute a Q function, representing the chance that a team scores the next goal, for {\em all} actions. 
\cite{Cervone2014a} applied a competing risk framework with Markov chain to model game context, and developed EPV, a point-wise conditional value similar to a Q function, for each action . The Q-function concept offers two key advantages for assigning values to actions \cite{schulte2017markov,decroos2018actions}: 1) All actions are scored on the same scale by looking ahead to expected outcomes. 2) Action values  reflect the match context in which they occur.  For example, a late check near the opponent's goal generates different scoring chances than a check at other locations and times. 

\begin{figure}[t!]
    \centering
    \includegraphics[width=0.45\textwidth] 
    {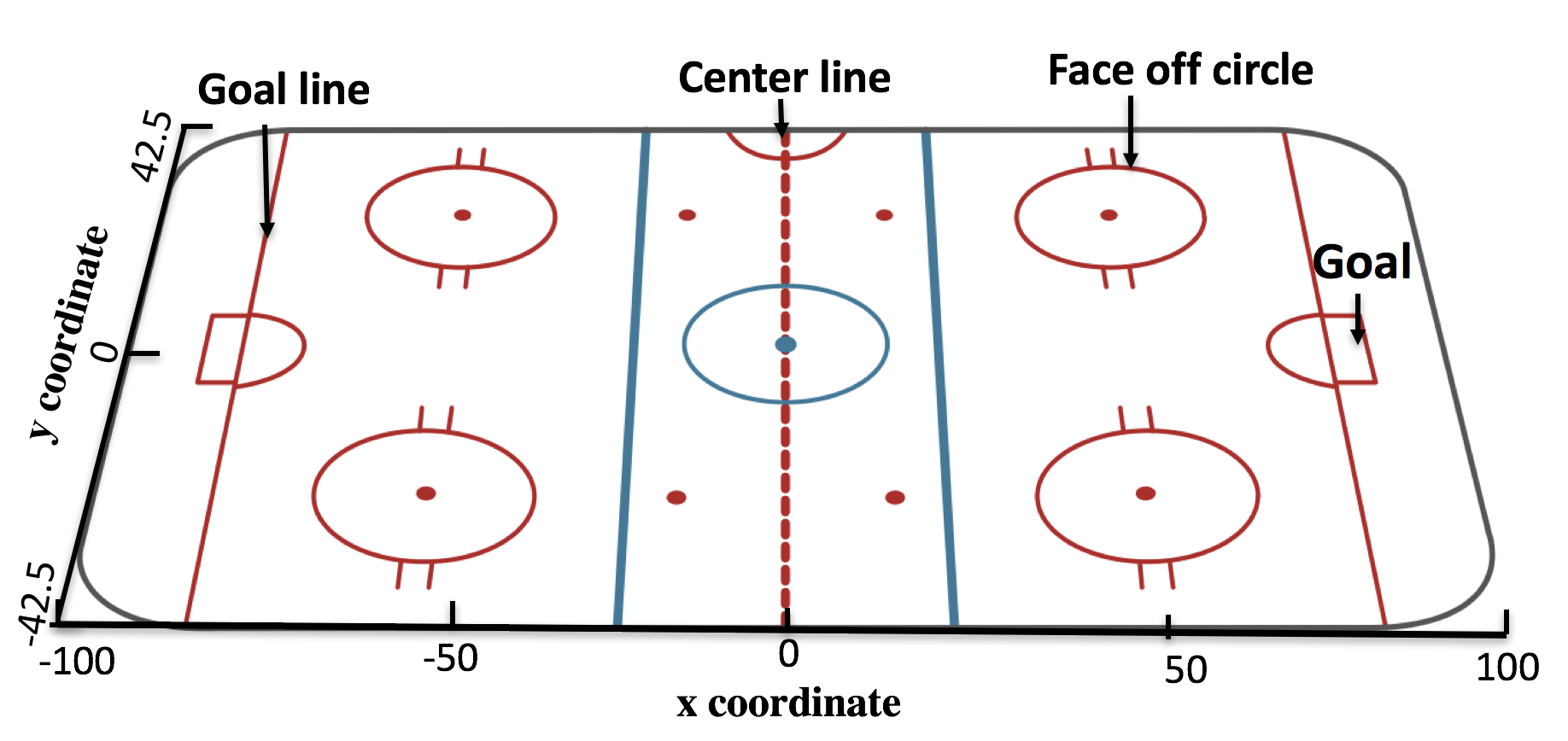}
    \caption{Ice Hockey Rink. \textbf{Ice hockey} is a fast-paced team sport, where two teams of skaters must shoot a puck into their opponent's net to score goals. 
    } 
    \label{fig:ice-hockey-rink}
\end{figure}

The states in the previous Markov models represent only a partial game context in the real sports match, but nonetheless the models assume full observability. 
Also, they pre-discretized input features, which leads to loss of information.
In this work, we utilize a deep reinforcement learning (DRL) model to learn an action-value Q function for capturing the current match context. The neural network representation can easily incorporate continuous quantities like rink location and game time. To handle partial observability, we introduce a possession-based Long Short Term Memory (LSTM) architecture that takes into account the current play history. 
Unlike most previous work on active reinforcement learning (RL), which aims to %directly 
compute {\em optimal strategies} for complex continuous-flow games~\cite{littlestone,Mnih2015}, we solve a prediction (not control) problem in the passive learning (on policy) setting \cite{Sutton1998}. {\em We use RL %prediction methods 
as a behavioral analytics tool for real human agents, not to control artificial agents.} 
 %\url{http://pages.cpsc.ucalgary.ca/~jacob/Courses/Winter2000/CPSC533/Slides/05.3-Reinforcement.pdf}}

Given a Q-function, the {\em impact} of an action is the change in Q-value due to the action. Our novel Goal Impact Metric (GIM)
% to evaluate a player's performance 
% is computed as follows. We define the {\em impact} of an action as the change in Q-value due to the action, and 
aggregates the impact of all actions of a player. To our knowledge, this is the first player evaluation metric based on DRL. The GIM metric measures both players' offensive and defensive contribution to goal scoring. For player evaluation, similar to clustering, ground truth is not available. A common methodology~\cite{Routley2015a,Pettigrew2015} is to assess the predictive value of a player evaluation metric for standard measures of success. 
Empirical comparison between 7 player evaluation metrics finds that 1) given a complete season, GIM correlates the most with 12 standard success measures and is the most temporally consistent metric, 2) given partial game information, GIM generalizes best to future salary and season total success.

\section{Related Work}
We discuss the previous work most related to our approach.

{\em Deep Reinforcement Learning.}  Previous DRL work has focused on {\em control} in continuous-flow games, not prediction~\cite{Mnih2015}. Among these papers, ~\cite{littlestone} use a very similar network architecture to ours, but with a fixed trace length parameter rather than our possession-based method. \citeauthor{littlestone} find that for partially observable control problems, the LSTM mechanism outperforms a  memory window. Our study confirms this finding in an on policy prediction problem. 

{\em Player Evaluation.} Albert et al. \citeyear{schwartz} provide several up-to-date survey articles about evaluating players. 
A fundamental difficulty for action value counts in continuous-flow games is that they traditionally have been restricted to goals and actions immediately related to goals (e.g. shots). The Q-function solves this problem by using lookahead to assign values to all actions.  

{\em  Player Evaluation with Reinforcement Learning.}
Using the Q-function to evaluate players is a recent development \cite{schulte2017markov,Cervone2014a,Routley2015a}. \citeauthor{schulte2017markov} discretized location and time coordinates and applied dynamic programming to learn a Q-function. Discretization leads to loss of information, undesirable spatio-temporal discontinuities in the Q-function, and generalizes poorly to unobserved parts of the state space. For basketball, \citeauthor{Cervone2014a} defined a player performance metric
%value-above-replacement metric 
based on an expected point value model that is equivalent to a Q-function. 
%\textcolor{orange}{They did not use reinforcement learning.} 
Their approach assumes complete observability (of all players at all times), while our data provide partial observability only. 

\section{Task Formulation and Approach} 
\begin{figure}[htb]
    \centering
    \includegraphics[width=0.50\textwidth] 
    {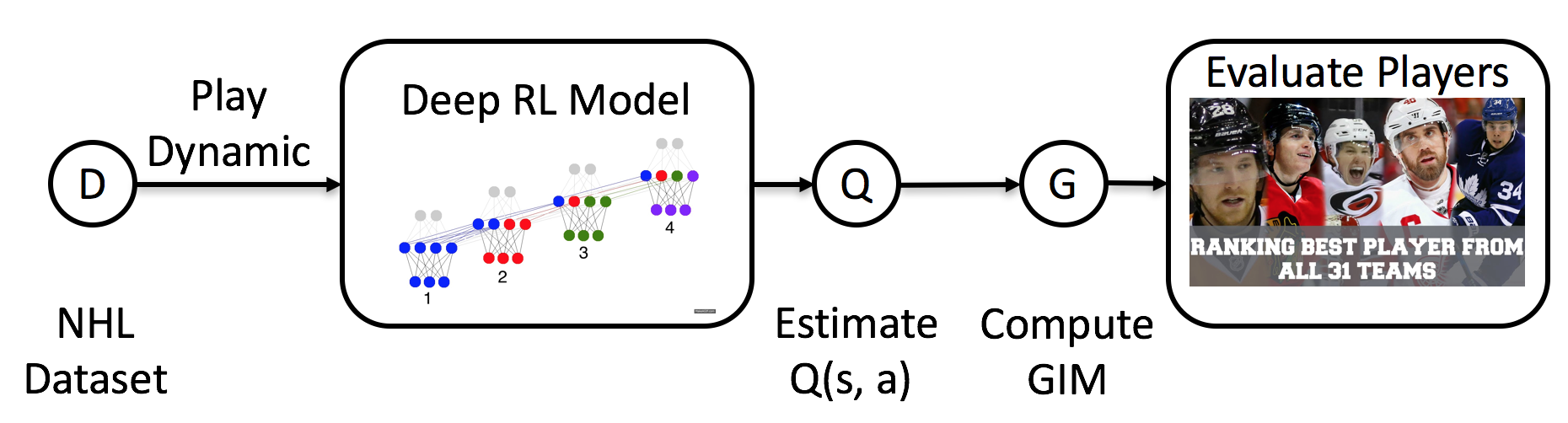}
    \caption{System Flow for Player Evaluation}
    \label{fig:control-flow}
\end{figure}

Player evaluation (the ``Moneyball'' problem) is one of the most studied tasks in sports analytics. %\cite[Ch.4]{Schumaker2010}. 
Players are rated by their observed performance over a set of games. Our approach to evaluating players is illustrated in Figure~\ref{fig:control-flow}. Given dynamic game tracking data, we 
apply Reinforcement Learning to estimate the {\em action value} function $\Qt{}{\history}{\action}$, which assigns a value to action $\action$ given game state $\history$. We define a new player evaluation metric called \textbf{Goal Impact Metric (GIM)} to value each player, based on the aggregated impact
% (difference between Q values) 
of their actions, which is defined in Section 6 below. 
Player evaluation is a descriptive task rather than a predictive generalization problem.% ~\cite[Ch.4]{Schumaker2010}. 
As game event data does not provide a ground truth rating of player performance, our experiments assess player evaluation as an unsupervised problem in Section 7.

\begin{table*}[htb!]
    \begin{center}
    {\footnotesize GID=GameId, PID=playerId, GT=GameTime, TID=TeamId, MP=Manpower, GD=Goal Difference, OC = Outcome, S=Succeed, F=Fail, P = Team Possess puck, H=Home, A=Away, H/A=Team who performs action, TR = Time Remain, PN = Play Number, D = Duration}
    \end{center}
    \begin{minipage}{.6\textwidth}
        % \caption{Dataset Example}
        % \label{table:example-of-dataset}
        \begin{center}
        \resizebox{\columnwidth}{!}{
        \begin{tabular}{|c|c|c|c|c|c|c|c|c|c|c|}
        \hline
        GID & PID & GT & TID & X & Y & MP & GD & Action & OC & P\\ \hline
        1365 & 126 & 14.3 & 6 & -11.0 & 25.5 & Even & 0 & Lpr & S & A\\ 
        1365 & 126 & 17.5 & 6 & -23.5 & -36.5 & Even & 0 & Carry & S & A\\
        1365 & 270 & 17.8 & 23 & 14.5 & 35.5 & Even & 0 & Block & S & A\\
        1365 & 126 & 17.8 & 6 & -18.5 & -37.0 & Even & 0 & Pass & F & A\\
        1365 & 609 & 19.3 & 23 & -28.0 & 25.5 & Even & 0 & Lpr & S & H\\
        1365 & 609 & 19.3 & 23 & -28.0 & 25.5 & Even & 0 & Pass & S & H\\ 
        \hline
        \end{tabular}
        }
        \end{center}
        \caption{Dataset Example}
        \label{table:example-of-dataset}
    \end{minipage}%
    \begin{minipage}{0.4\textwidth}
        % \caption{Derived Features}
        % \label{table:derived-features}
        \begin{center}
        \resizebox{\columnwidth}{!}{
        \begin{tabular}{|c|c|c|c|c|c|}
        \hline
        Velocity & TR & D & Angle & H/A & PN \\ 
        \hline
        (-23.4, 1.5)   & 3585.7 & 3.4 & 0.250 & A & 4  \\ 
        (-4.0, -3.5)   & 3582.5 & 3.1 & 0.314 & A & 4 \\
        (-27.0, -3.0)   & 3582.2 & 0.3 & 0.445 & H & 4\\
        (0, 0)         & 3582.2 & 0.0 & 0.331 & A & 4\\
        (-30.3, -7.5)   & 3580.6 & 1.5 & 0.214 & H & 5\\
        (0,0)          & 3580.6 & 0.0 & 0.214 & H & 5\\ 
        \hline
        \end{tabular}
        }
        \end{center}
        \caption{Derived Features}
        \label{table:derived-features}
    \end{minipage}
    % \begin{center}
    % {\footnotesize GID=GameId, PID=playerId, GT=GameTime, TID=TeamId, MP=Manpower, GD=Goal Difference, OC = Outcome, S=Succeed, F=Fail, P = Team Possess puck, H=Home, A=Away, H/A=Team who performs action, TR = Time Remain, PN = Play Number, D = Duration}
    % \end{center}
\end{table*}

\begin{table}[htb]
% \caption{Complete Feature List}
% \label{table:feature-of-dataset}
\begin{center}
\resizebox{\columnwidth}{!}{
\begin{tabular}{|c|c|c|}
\hline
\bf{Name} & \bf{Type} & \bf{Range} \\ \hline
X Coordinate of Puck & Continuous & [-100, 100]\\
Y Coordinate of Puck & Continuous & [-42.5, 42.5]\\
Velocity of Puck & Continuous & (-inf, +inf)\\
Game Time Remain & Continuous & [0, 3600]\\
Score Differential & Discrete & (-inf, +inf)\\
Manpower Situation & Discrete & \{EV, SH, PP\}\\
Event Duration& Continuous & [0, +inf) \\
Action Outcome & Discrete & \{successful, failure\}\\
Angle between puck and goal & Continuous & [$-3.14$, $3.14$]\\
Home or Away Team & Discrete & \{Home, Away\} \\ \hline
\end{tabular}
}
\end{center}
\caption{Complete Feature List}
\label{table:feature-of-dataset}
\end{table} 

\section{Play Dynamic in NHL}

% rank players by state action Q-function.

% \textcolor{red}{You can copy this from the MIT Sloan paper. I would suggest adding a sample of a data table like we did in the Sloan paper to make the data vivid.}
We utilize a \textbf{dataset} constructed by SPORTLOGiQ using computer vision techniques. 
% including player tracking and activity recognition. 
% It consists of play-by-play information of game events and player actions for the entire 2015-2016 NHL season. 
The data provide information about {\em game events} and {\em player actions} for the entire 2015-2016 NHL (largest professional ice hockey league) season,
which contains 3,382,129 events, covering 30 teams, 1140 games and 2,233 players. 
% \textcolor{red}{it would be better to have the dataset stats in a table, see UAI paper} \textcolor{blue}{I am worrying about space, let's see if we still have enough space here.} 
Table \ref{table:example-of-dataset} shows an excerpt. 
% A breakdown of this dataset is shown in table \ref{table:size-of-dataset}. 
% \begin{table}[htb]
% \caption{Dataset Statistics.}
% \label{table:size-of-dataset}
% \begin{center}
% \begin{tabular}{|l|c|}
% \hline
% \bf{Number of Teams} & 30 \\ \hline
% \bf{Number of Players} & 2,233 \\ \hline
% \bf{Number of Games} & 1,140 \\ \hline
% \bf{Number of Events} & 3,382,129 \\ \hline
% \end{tabular}
% \end{center}
% \end{table}
The data track events around the puck, and record the identity and actions of the player in possession, with space and time stamps, and features of the game context. 
% We used 13 of the recorded action types listed in the supplementary material.
The table utilizes adjusted spatial coordinates where negative numbers refer to the defensive zone of the acting player, positive numbers to his offensive zone. Adjusted X-coordinates run from -100 to +100, Y-coordinates from 42.5 to -42.5, and the origin is at the ice center as in Figure~\ref{fig:ice-hockey-rink}. We augment the data with derived features in Table \ref{table:derived-features} 
and list the complete feature set in Table \ref{table:feature-of-dataset}.

%\textcolor{blue}{Which one is better?  (1) We construct a model of NHL play dynamics from tracking data.  (2) We apply Markov Game Framework~\cite{Littman1994} to model ice hockey dynamics.} %we use model-free so neither is good
We apply the Markov Game framework~\cite{Littman1994} to learn an action value function for NHL play.
Our notation for RL concepts follows \cite{Mnih2015}. 
% We treat players as agents.
%\textcolor{blue}{how about agent?} 
There are two agents $\home$ resp. $\away$ representing the home resp. away team. The \defterm{reward}, represented by goal vector $\goal_t$ is a 1-of-3 indicator vector that specifies which team scores ($\home,\away,\none$). 
%For readability, we use $\home,\away,\none$ to denote the team in a goal vector. For example, $\goal_{t,\home}=1$ means that the home team scores at time $t$. 
% , and for each goal indicator.
An \defterm{action} $\action_t$ is one of 13 types, including shot, block, assist, etc.,
%listed in Table~\ref{table:action-types}\textcolor{red}{do you mean supplementary material}, 
together with a mark that specifies the team executing the action, e.g. $\it{Shot}(\home)$. 
%), the away team scores ($\goal_{t,2} = 1$), or neither team scores ($\goal_{t,3}=1$).
%
% \textcolor{blue}{value for home and away actually are indenpedent, even if we train them together. home team scores ($\goal=[1, 0]$), the away team scores ($\goal=[0,  1]$), or neither team scores ($\goal=[0, 0]$}. \textcolor{red}{I incorporated this idea when defining the value function. This is just meant to define simple labels for describing the data.}
An \defterm{observation} is a feature vector $\features_{t}$ for discrete time step $t$ that specifies a value for the 10 features listed in Table~\ref{table:feature-of-dataset}. We use the complete sequence $\history_{t} \equiv (\features_t,\action_{t-1},\features_{t-1},\ldots,\features_0)$ as the state representation at time step $t$ \cite{Mnih2015}, which satisfies the Markov property.

We divide NHL games into {\bf goal-scoring episodes}, so that each episode 1) begins at the beginning of the game, or immediately after a goal, and 2) terminates with a goal or the end of the game. 
A \defterm{$Q$ function} represents the conditional probability of the event that the home resp. away team {\em scores the goal at the end of the current  episode} (denoted $\egoal_{\home}=1$ resp. $\egoal_{\away}=1$), or neither team does (denoted $\egoal_{\none}=1$):

$$ Q_{\team}(\history,\action) = P(\egoal_{\team}=1|\history_{t}=\history,\action_{t}=\action)$$

where $\team$ is a placeholder for one of $\home,\away,\none$. This $Q$-function represents {\em the probability that a team scores the next goal}, given current play dynamics in the NHL (cf. \citeauthor{schulte2017markov,Routley2015a}). Different $Q$-functions for different expected outcomes have been used to capture different aspects of NHL play dynamics, such as match win \cite{Pettigrew2015,kaplan,Routley2015a} and penalties \cite{Routley2015a}. 
For player evaluation, the next-goal Q function has three advantages.
1) The next-goal reward captures what a coach expects from a player. For example, if a team is ahead by two goals with one minute left in the match, a player's actions have negligible effect on final match outcome. Nonetheless professionals should keep playing as well as they can and maximize the scoring chances for their own team. 2) %As we illustrate below, 
The $Q$-values are easy to interpret, since they model the probability of an event that is a relatively short time away (compared to final match outcome). 3) Increasing the probability that a player's team scores the next goal captures both offensive and defensive value. For example, a defensive action like blocking a shot decreases the probability that the other team will score the next goal, thereby increasing the probability that the player's own team will score the next goal.

\begin{figure}[htb]
    \centering
    \includegraphics[width=0.5\textwidth] 
    {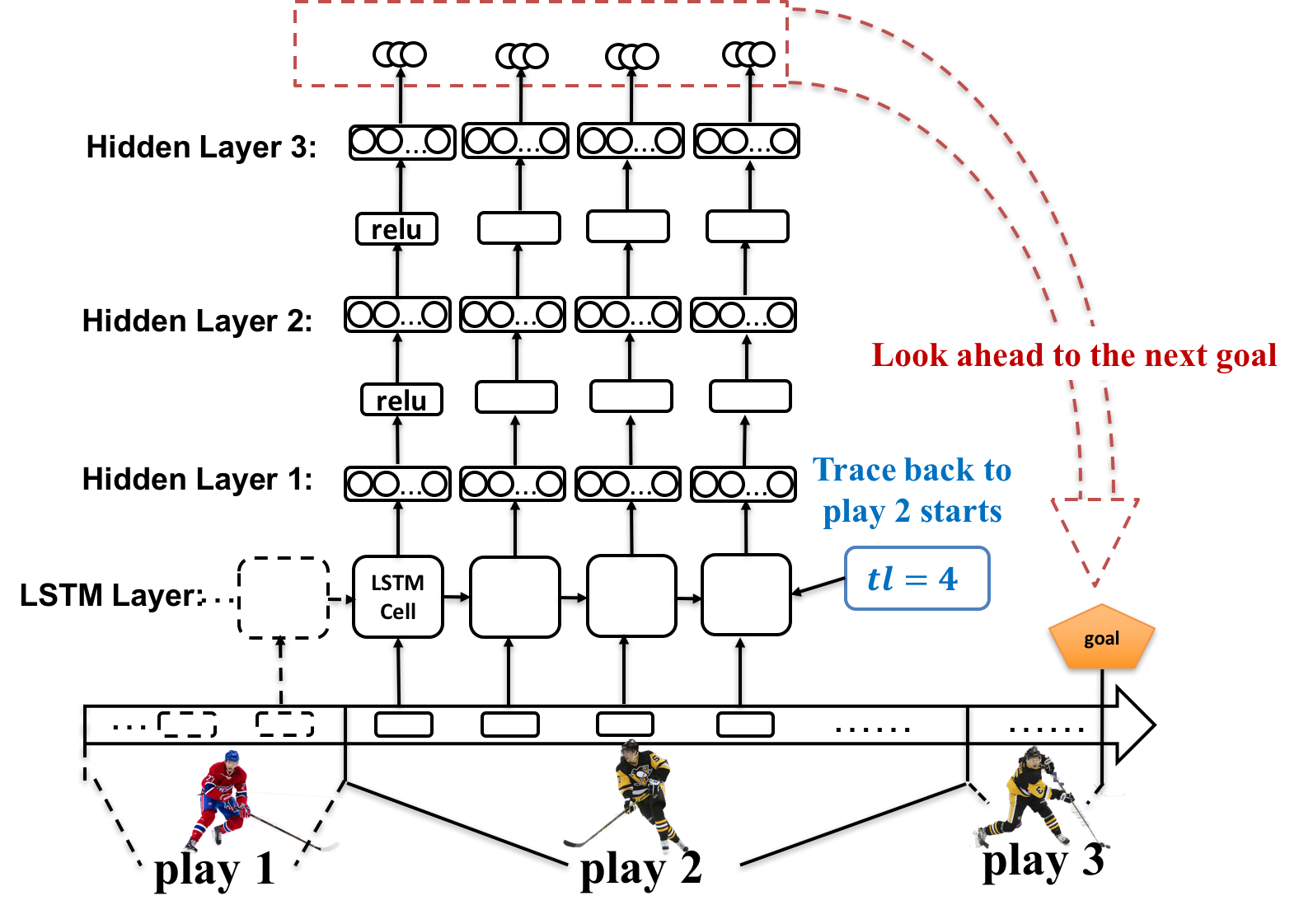}
    \caption{Our design is a 5-layer network with 3 hidden layers. Each hidden layer contains 1000 nodes, which utilize a relu activation function. The first hidden layer is the LSTM layer, the remaining layers are fully connected. Temporal-difference learning looks ahead to the next goal, and the LSTM memory traces back to the beginning of the play (the last possession change).}
    \label{fig:DP-look-trace}
\end{figure}

\section{Learning Q values with DP-LSTM Sarsa} 
% \textcolor{blue}{According to first reviewer's opinion, to avoid confusion, the authors could be more precise by calling their algorithm Sarsa-value-prediction.} 
We take a function approximation approach and learn a neural network that represents the $Q$-function~ ($Q_{\team}(\history,\action)$). 

\subsection{Network Architecture} Figure~\ref{fig:DP-look-trace} shows our model structure. Three output nodes represent the estimates $\Qmodel{\home}{\history}{\action}$, $\Qmodel{\away}{\history}{\action}$ and $\Qmodel{\none}{\history}{\action}$. Output values are normalized to  probabilities. The $\hat{Q}$-functions for each team share weights. The network architecture is a Dynamic LSTM  that takes as inputs a current sequence $\history_{t}$, an action $\action_{t}$ and a dynamic trace length $\tracel_{t}$.\footnote{We experimented with a single-hidden layer, but weight training failed to converge.}  %, and is trained to predicted the expected returns. 

\begin{figure}[htb]
    \centering
    \includegraphics[width=0.5\textwidth] 
    {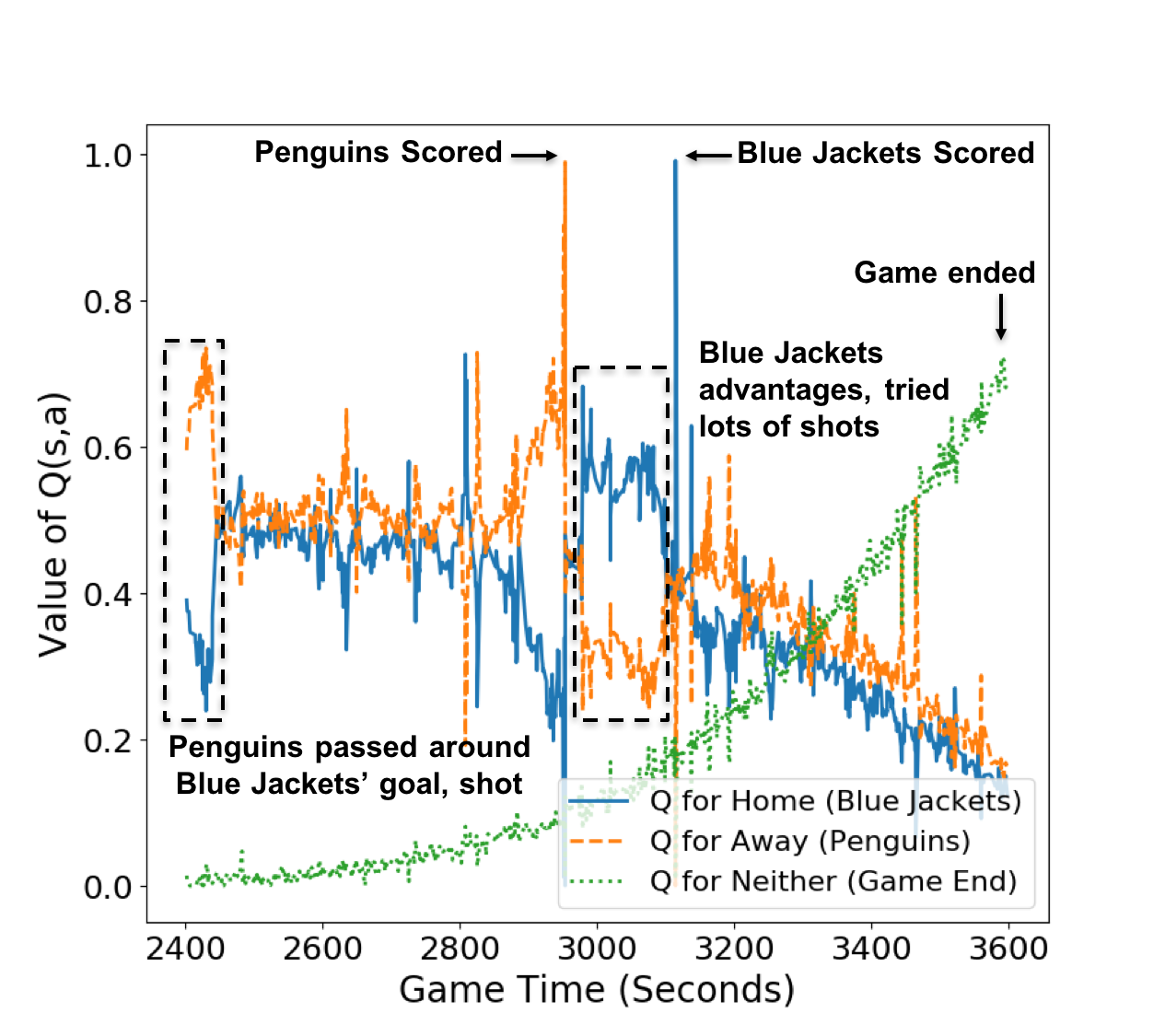}
    \caption{Temporal Projection of the method. For each team, and each game time, the graph shows the chance the that team scores the next goal, as estimated by the model. Major events lead to major changes in scoring chances, as annotated. The network also captures smaller changes associated with {\em every} action under different game contexts. 
%     \textcolor{red}{even better would be showing game frames. Like in the Human-level control paper. Probably something for later or maybe the conference presentation.}\textcolor{blue}{?? need to discuss}
    }
    \label{fig:value-ticker}
\end{figure}
\subsection{Weight Training}

%In a game setting, the $Q$ function depends on a policy for both the home and the away team, denoted $\pi_{\home},\pi_{\away}$, which describe the behaviour of a generic home resp. away team in the NHL. 
We apply an on-policy Temporal Difference (TD) prediction method \defterm{Sarsa}~\cite[Ch.6.4]{Sutton1998}, to estimate $Q_{\team}(\history,\action)$ for the NLH play dynamics observed in our dataset. 
%current policies $\pi_{\home}$, $\pi_{\away}$, etc.\todo{why "etc."}. 
Weights $\theta$
% \textcolor{red}{should we use vector notation?} \textcolor{blue}{do you mean $\Vec{\theta_{t}}$} \textcolor{red}{yes but these days people usually use bold like this $\bs{\theta}$} 
are optimized by minibatch gradient descent via backpropagation. We used batch size 32 (determined experimentally). The Sarsa gradient descent update at time step $t$ is based on  a squared-error loss function:

% \begin{align*}
% \mathcal{L}_{t}(\theta_{t}) & =  \Expect_{\history_t, \action_t, \goal_{t}, \history_{t+1}}[(\goal_{t} + \Qmodel{}{\history_{t+1}}{\action_{t+1}, \theta_t} - \Qmodel{}{\history_t}{\action_t, \theta_t})^{2}] \\
% \theta_{t+1} & =  \theta_{t} + \alpha \nabla_{\theta} \mathcal{L}(\theta_{t})
% \end{align*}

\begin{align*}
\mathcal{L}_{t}(\theta_t) & =  \Expect[(\goal_{t} + \Qmodel{}{\history_{t+1}}{\action_{t+1}, \theta_t} - \Qmodel{}{\history_t}{\action_t, \theta_t})^{2}] \\
\theta_{t+1} & =  \theta_{t} + \alpha \nabla_{\theta} \mathcal{L}(\theta_{t})
\end{align*}

where $\goal$ and $\hat{Q}$ are for a single team. 
%For {\em data preprocessing,} we standardize input features to have zero mean and unit variance.   
LSTM training requires setting a {\em trace length} $\tracel_{t}$ parameter. 
This key parameter controls how far back in time the LSTM propagates the error signal from the current time at the input history.
%\textcolor{blue}{
 %we restrict history traces to remain within the same offensive play.}
Team sports like Ice Hockey show a turn-taking aspect where one team is on the offensive and the other defends; one such turn is called a {\em play}. 
We set $\tracel_{t}$ to the number of time steps from current time $t$ to the beginning of the current {\em play} (with a maximum of 10 steps). A play ends when the possession of puck changes from one team to another.
% {\em Plays} are constructed by breaking a {\em goal-scoring episode} to sub-episodes \textcolor{blue}{reviewer 2 said two episodes confuse him, can we change its name?} when the possession of puck changes from one team to another, as  using possession changes to define episodes is common in several continuous-flow sports, especially basketball \cite{Cervone2014a,omidiran2011new}. 
Using possession changes as break points for temporal models %to define episodes 
is common in several continuous-flow sports, especially basketball \cite{Cervone2014a,omidiran2011new}. We apply Tensorflow to implement training;  our source code is published on-line.\footnote{https://github.com/Guiliang/DRL-ice-hockey}
% to allow others to experiment with our model. \textcolor{blue}{I will do it}

\emph{Illustration of Temporal Projection.} 
Figure~\ref{fig:value-ticker} shows a value ticker \cite{Decroos2017,Cervone2014a} that represents the evolution of the Q function from the $3^{rd}$ period of a match between the Blue Jackets (Home team) and the Penguins (Away team), Nov. 17, 2015. The figure plots values of the three output nodes. We highlight critical events and match contexts to show the context-sensitivity of the Q function. High scoring probabilities for one team decrease those of its opponent. The probability that neither team scores rises significantly at the end of the match.

\section{Player Evaluation} In this section, we define our novel Goal Impact Metric and give an example player ranking.

\subsection{Player Evaluation Metric}

Our $Q$-function concept provides a novel AI-based definition for assigning a value to an action. 
% \paragraph{Total Action Values Based on the Q-function}
Like~\cite{schulteapples}, we measure the quality of an action by how much it changes the expected return of a player's team. Whereas the scoring chance at a time measures the value of a state, and therefore depends on the previous efforts of the entire team, the change in value measures directly the impact of an action by a specific player. In terms of the Q-function, this is the {\em change in Q-value} due to a player's action. This quantity is defined as the action's \defterm{impact}. 
The impact can be visualized as the difference between successive points in the Q-value ticker (Figure~\ref{fig:value-ticker}). 
For our specific choice of Next Goal as the reward function, we refer to \defterm{goal impact}.
% \textcolor{red}{Goal impact captures a player's offensive contributions---increasing the scoring chances of his team---as well as his defensive contributions---decreasing the scoring chances of the opposing team. - we kind of said this earlier} 
The total impact of a player's actions is his \defterm{Goal Impact Metric} (GIM). The formal equations are:

\begin{align*}
\impact{\team}{\history_{t},\action_{t}} & = \Qt{\team}{\history_{t}}{\action_{t}}-\Qt{\team}{\history_{t-1}}{\action_{t-1}} \label{eq:impact} \\
GIM^{i}(\sequences) & = \sum_{\history,\action} \pcounts{i}{\history}{\action}{\sequences} \times \impact{\team_{i}}{\history,\action}
\end{align*}

where $\sequences$ indicates our dataset, $\team_{i}$ denotes the team of player $i$, and $\pcounts{i}{\history}{\action}{\sequences}$ is the number of times that player $i$ was observed to perform action $\action$ at $\history$. Because it is the sum of differences between subsequent Q values, the GIM metric inherits context-sensitivity from the Q function. 
%\textcolor{blue}{GIM is the sum of difference between context aware Q-functions which measures action's influence to game context. So GIM is context-aware.}

\subsection{Rank Players with GIM}

Table~\ref{table:top-20-ranking} lists the top-20 highest impacts players, with basic statistics. All these players are well-known NHL stars.  Taylor Hall tops the ranking although he did not score the most goals. This shows how our ranking, while correlated with goals, also {\em reflects the value of other actions by the player.} For instance, we find that the total number of passes performed by Taylor Hall is exceptionally high at 320.
% For instance, our ranking reflects that the total number of Hall's passes is exceptionally high at 320.
% \textcolor{blue}{how about 'For instance, we find the total number of passes performed by Taylor Hall, who ranked highest with GIM, is exceptionally high at 320.'}
Our metric can be used to {\em identify undervalued players.} For instance, Johnny Gaudreau and Mark Scheifele drew salaries below what their GIM rank would suggest. Later they received a $\$5M+ $ contract for the 2016-17 season.

\begin{table}[htb]
% \caption{2015-2016 Top-20 Player Impact Scores 
% % \textcolor{blue}{do we really need to include +/- here?}
% }
% \label{table:top-20-ranking}
\begin{center}
\resizebox{\columnwidth}{!}{
\begin{tabular}{ccccccc}
\hline
Name      & GIM & Assists & Goals & Points & Team & Salary \\ \hline
Taylor Hall        & 96.40           & 39               & 26             & 65              & EDM           & \$6,000,000     \\
Joe Pavelski       & 94.56           & 40               & 38             & 78              & SJS           & \$6,000,000     \\
Johnny Gaudreau    & 94.51           & 48               & 30             & 78              & CGY            & \$925,000       \\
Anze Kopitar       & 94.10           & 49               & 25             & 74              & LAK           & \$7,700,000     \\
Erik Karlsson      & 92.41           & 66               & 16             & 82              & OTT           & \$7,000,000     \\
Patrice Bergeron   & 92.06           & 36               & 32             & 68              & BOS           & \$8,750,000     \\
Mark Scheifele     & 90.67           & 32               & 29             & 61              & WPG           & \$832,500       \\
Sidney Crosby      & 90.21           & 49               & 36             & 85              & PIT           & \$12,000,000    \\
Claude Giroux      & 89.64           & 45               & 22             & 67              & PHI          & \$9,000,000     \\
Dustin Byfuglien   & 89.46           & 34               & 19             & 53              & WPG            & \$6,000,000     \\
Jamie Benn         & 88.38           & 48               & 41             & 89              & DAL            & \$5,750,000     \\
Patrick Kane       & 87.81           & 60               & 46             & 106             & CHI           & \$13,800,000    \\
Mark Stone         & 86.42           & 38               & 23             & 61              & OTT           & \$2,250,000     \\
Blake Wheeler      & 85.83           & 52               & 26             & 78              & WPG            & \$5,800,000     \\
Tyler Toffoli      & 83.25           & 27               & 31             & 58              & DAL           & \$2,600,000     \\
Charlie Coyle      & 81.50           & 21               & 21             & 42              & MIN            & \$1,900,000     \\
Tyson Barrie       & 81.46           & 36               & 13             & 49              & COL          & \$3,200,000     \\
Jonathan Toews     & 80.92           & 30               & 28             & 58              & CHI           & \$13,800,000    \\
Sean Monahan       & 80.92           & 36               & 27             & 63              & CGY           & \$925,000       \\
Vladimir Tarasenko & 80.68           & 34               & 40             & 74              & STL            & \$8,000,000     \\ \hline
\end{tabular}
}
\end{center}
\caption{2015-2016 Top-20 Player Impact Scores 
% \textcolor{blue}{do we really need to include +/- here?}
}
\label{table:top-20-ranking}
\end{table}

\section{Empirical Evaluation}

We describe our comparison methods and evaluation methodology. Similar to clustering 
%and recommendation 
problems, there is {\em no ground truth} for the task of player evaluation. To assess a player evaluation metric, we follow previous work \cite{Routley2015a,Pettigrew2015} and compute its correlation with  statistics that directly measure success like Goals, Assists, Points, Play Time (Section 7.2). There are two justifications for comparing with  {\em success measures}. (1) These statistics are generally recognized as important measures of a player's strength, because they indicate 
%recognition, in the case of salary, and 
the player's ability to contribute to %important 
game-changing events. So a comprehensive  performance metric ought to be related to them. (2) The success measures are often forecasting targets for hockey stakeholders, so a good player evaluation metric should have predictive value for them. For example, teams would want to know how many points an offensive player will contribute. 
% Points are also the metric used to score players in fantasy play, which engages thousands of fans and is strongly supported by the NHL.
% \textcolor{orange}{
% In hockey fantasy leagues, fans assemble their own virtual teams, which are then scored on their total points in the actual NHL play. Fantasy play engages thousands of fans and is extensively supported by the NHL.}\footnote{\url{ https://www.nhl.com/news/2016-17-left-wing-fantasy-rankings/c-281561644}}. 
 To evaluate the ability of the GIM metric for generalizing from past performance to future success, we report two measurements: How well the GIM metric predicts a total season success measure %when it is computed 
from a sample of matches only (Section 7.3), and how well the GIM metric predicts the future salary of a player in subsequent seasons (Section 7.4). 
%Predicting a player's salary has been studied in sports analytics because it supports salary negotiations between players and teams~\cite{idson2000team}.
Mapping performance to salaries is a practically important task because it provides an objective standard to guide players and teams in salary negotiations \cite{idson2000team}.

\subsection{Comparison Player Evaluation Metrics} \label{sec:metrics}

We compare GIM with the following player evaluation metrics to show the advantage of 1) modeling game context 2) incorporating continuous context signal 3) including history.

% \begin{description}
Our first baseline method \textbf{Plus-Minus (+/-)} is a commonly used metric that measures how the presence of a player influences the goals of his team~\cite{Macdonald2011}.
The second baseline method \textbf{Goal-Above-Replacement (GAR)} estimates the difference of team's scoring chances when the target player plays, vs. replacing him or her with an average player~\cite{gerstenberg2014wins}.
\textbf{Win-Above-Replacement (WAR)}, our third baseline method, is the same as GAR but for winning chances~\cite{gerstenberg2014wins}.
Our fourth baseline method \textbf{Expected Goal (EG)} weights each shot by the chance of it leading to a goal.
These four methods consider only very limited game context.
The last baseline method \textbf{Scoring Impact (SI)} is the most similar method to GIM based on Q-values. But Q-values are learned with pre-discretized spatial regions and game time~\cite{schulte2017markov}.
As a lesion method, we include \textbf{GIM-T1}, where we set the maximum trace length of LSTM to 1 (instead of 10) in computing GIM. This comparison assesses the importance of including enough history information.
% \end{description}

{\em Computing Cost.} Compared to traditional metrics like +/-, learning a Q-function is computationally demanding (over 5 million gradient descent steps on our dataset). However, after the model has been trained off-line, the GIM metric can be computed quickly with a single pass over the data.
%\textcolor{blue}{Our training took 5 days computation time and over 5 million minibatch gradient descent steps.} 

{\em Significance Test.} To assess whether GIM is significantly different from the other player evaluation metrics, 
we perform paired t-tests over all players.  The null hypothesis is rejected with respective p-values: $1.1*10^{-186}$, $7.6*10^{-204}$, $8*10^{-218}$, $3.9*10^{-181}$, $4.7*10^{-201}$ and $1.3*10^{-05}$ for PlusMinus, GAR, WAR, EG, SI and GIM-T1, which shows that GIM values are very different from other metrics' values.

\subsection{Season Totals: Correlations with standard Success Measures} \label{sec:season-total}

In the following experiment, we compute the correlation between player ranking metrics and success measures over the entire season. 
Table~\ref{table:all-correlation} shows the correlation coefficients of the comparison methods with 14 standard success measures:  Assist, Goal, Game Wining Goal (GWG), Overtime Goal (OTG), Short-handed Goal (SHG), Power-play Goal (PPG), Shots (S), Point, Short-handed Point (SHP), Power-play Point (PPP), Face-off Win Percentage (FOW), Points Per Game (P/GP), Time On Ice (TOI) and Penalty Minute (PIM). These are all commonly used measures available from the NHL official website (www.nhl.com/stats/player).
{\em GIM achieves the highest correlation in 12 out of 14 success measures.} For the remaining two (TOI and PIM), GIM is comparable to the highest. Together, the Q-based metrics GIM, GIM-1 and SI show the highest correlations with success measures. EG is only the fourth best metric, because it considers only the expected value of shots without look-ahead. 
The traditional sports analytics metrics correlate poorly with almost all success measures. This is evidence that AI techniques that provide fine-grained expected action value estimates lead to better performance metrics. 
% than traditional sports analytics metrics. 
With the neural network model, GIM can handle continuous input without pre-discretization. This prevents the loss of game context information and explains why both GIM and GIM-T1 performs better than SI in most success measures. And the higher correlation of GIM compared to GIM-T1 also demonstrates the value of game history. In terms of absolute correlations, GIM achieves high values, except for the very rare events OTG, SHG,  SHP and FOW. Another exception is Penalty Minutes (PIM), which interestingly, show positive correlation with all player evaluation metrics, although penalties are undesirable. We hypothesize that better players are more likely to receive penalties, because they play more often and more aggressively. 
%and are more aggressive on court. 

\begin{table}[htb!]
\centering
% \caption{Correlation with standard success measures.
% %\textcolor{blue}{Time-on-ice Per Game (TOI) has been changed to Total Play Time (TPT) as your suggestion}
% }
% \label{table:all-correlation}
\resizebox{1\columnwidth}{!}{
\begin{tabular}{cccccccc}
\hline
methods & Assist & Goal & GWG & OTG & SHG & PPG & S \\ \hline
+/- & 0.236 & 0.204 & 0.217 & 0.16 & 0.095 & 0.099 & 0.118 \\
GAR & 0.527 & 0.633 & 0.552 & 0.324 & 0.191 & 0.583 & 0.549 \\
WAR & 0.516 & 0.652 & 0.551 & 0.332 & 0.192 & 0.564 & 0.532 \\ \hline
EG & 0.783 & 0.834 & 0.704 & 0.448 & 0.249 & 0.684 & 0.891 \\
SI & 0.869 & 0.745 & 0.631 & 0.411 & 0.27 & 0.591 & 0.898 \\
GIM-T1 & 0.873 & 0.752 & 0.682 & 0.428 & 0.291 & 0.607 & 0.877 \\
\textbf{GIM} & \textbf{0.875} & \textbf{0.878} & \textbf{0.751} & \textbf{0.465} & \textbf{0.345} & \textbf{0.71} & \textbf{0.912} \\ \hline
\\
\end{tabular}
}
\resizebox{1\columnwidth}{!}{
\begin{tabular}{cccccccc}
\hline
methods & Point & SHP & PPP & FOW & P/GP & TOI & PIM \\ \hline
+/- & 0.237 & 0.159 & 0.089 & -0.045 & 0.238 & 0.141 & 0.049 \\
GAR & 0.622 & 0.226 & 0.532 & 0.16 & 0.616 & 0.323 & 0.089 \\
WAR & 0.612 & 0.235 & 0.531 & 0.153 & 0.605 & 0.331 & 0.078 \\ \hline
EG & 0.854 & 0.287 & 0.729 & 0.28 & 0.702 & 0.722 & 0.354 \\
SI & 0.869 & 0.37 & 0.707 & 0.185 & 0.655 & \textbf{0.955} & \textbf{0.492} \\
GIM-T1 & 0.902 & 0.384 & 0.736 & 0.288 & 0.738 & 0.777 & 0.347 \\
\textbf{GIM} & \textbf{0.93} & \textbf{0.399} & \textbf{0.774} & \textbf{0.295} & \textbf{0.749} & 0.835 & 0.405 \\ \hline
\end{tabular}
}
\caption{Correlation with standard success measures.
%\textcolor{blue}{Time-on-ice Per Game (TOI) has been changed to Total Play Time (TPT) as your suggestion}
}
\label{table:all-correlation}
\end{table}

\subsection{Round-by-Round Correlations: Predicting Future Performance From Past Performance} \label{sec:round-by-round}

A sports season is commonly divided into \textbf {rounds}. In round $n$, a team or player  has finished $n$ games in a season. For a given performance metric, we measure the correlation between (i) its value computed {\em over the first $n$ rounds}, and (ii) the value of the three main success measures, assists, goals, and points, computed {\em over the entire season}. This allows us to assess how quickly different metrics acquire predictive power for the final season total, so that future performance can be predicted from past performance. We also evaluate the {\em auto-correlation} of a metric's round-by-round total with its own season total. The auto-correlation is a measure of temporal consistency, which is a desirable feature~\cite{Pettigrew2015}, because generally the skill of a player does not change greatly throughout a season. Therefore a good performance metric should show temporal consistency. 

\begin{figure}[htb]

\begin{minipage}{.5\columnwidth}
\centering\
\subfloat{\label{main:a}\includegraphics[scale=.22]{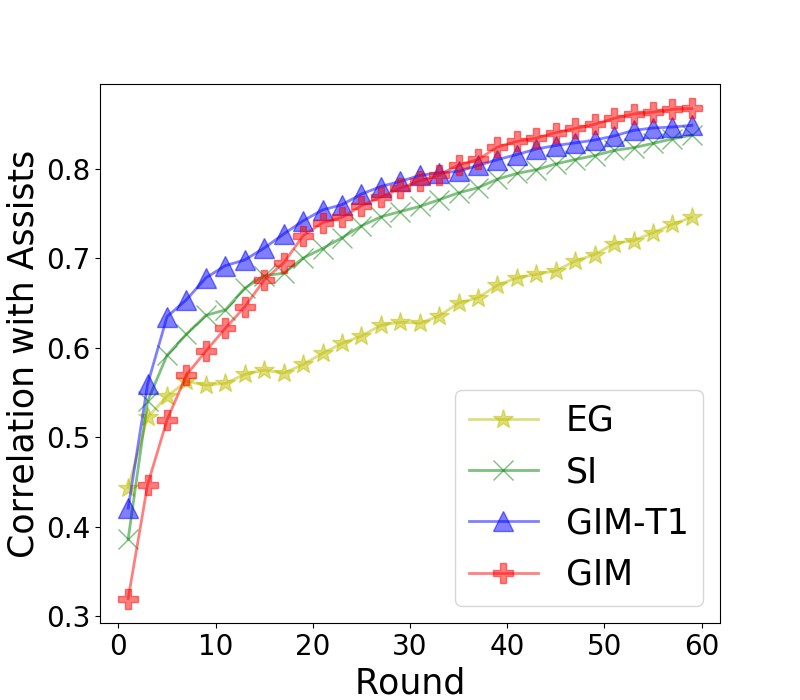}}
\end{minipage}%
\begin{minipage}{.5\columnwidth}
\centering
\subfloat{\label{main:b}\includegraphics[scale=.22]{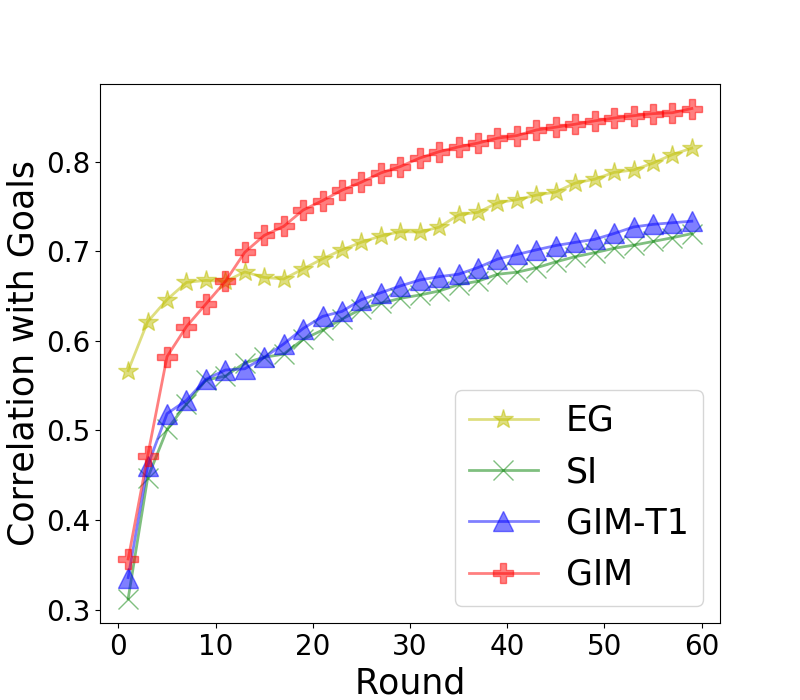}}
\end{minipage}\par\smallskip
\begin{minipage}{.5\columnwidth}
\centering
\subfloat{\label{main:c}\includegraphics[scale=.22]{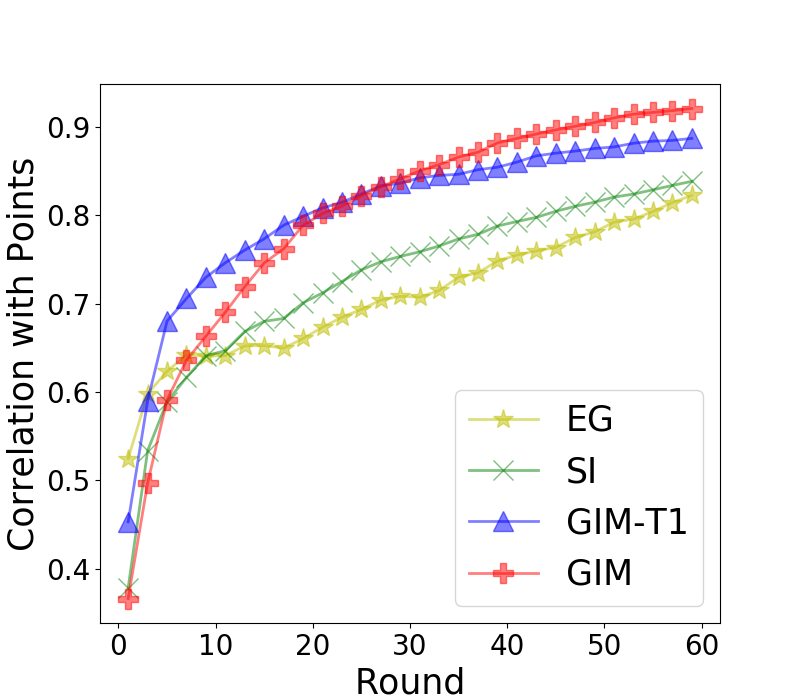}}
\end{minipage}%
\begin{minipage}{.5\columnwidth}
\centering
\subfloat{\label{main:d}\includegraphics[scale=.22]{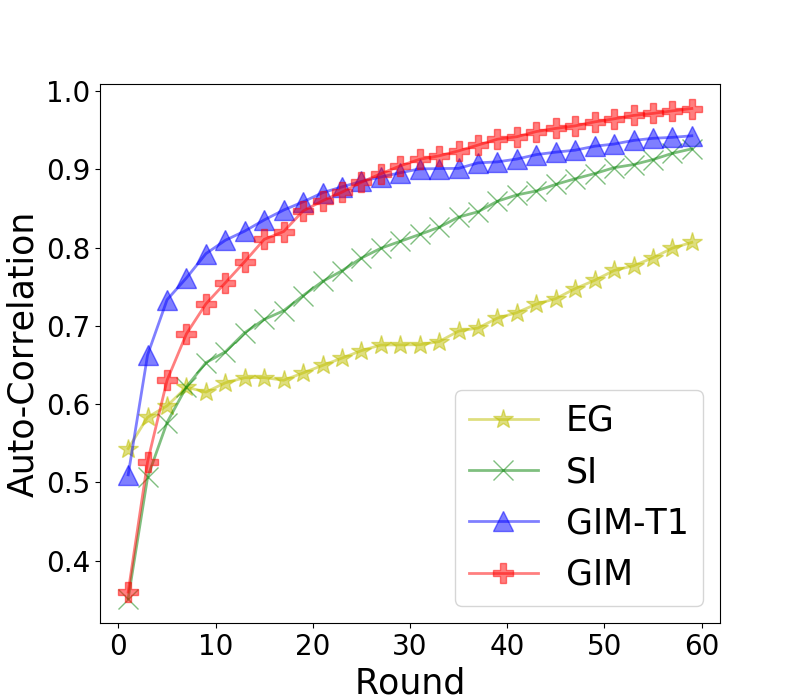}}
\end{minipage}
\caption{Correlations between round-by-round metrics and season totals.
% \textcolor{blue}{I will complement experiment with GIM-t1 and GIM-t5}
}
\label{fig:round-by-round-correlation}
\end{figure}

We focused on the expected value metrics EG, SI, GIM-T1 and GIM, which had the highest correlations with success in Table~\ref{table:all-correlation}.
Figure~\ref{fig:round-by-round-correlation} shows metrics' round-by-round correlation coefficients with assists, goals, and points. 
The bottom right shows the auto-correlation of a metric's round-by-round total with its own season total.  {\em GIM is the most stable metric} as measured by auto-correlation: after half the season, the correlation between the round-by-round GIM and the final GIM is already above 0.9.

We find both GIM and GIM-T1 eventually dominate the predictive value of the other metrics, which shows the advantages of modeling sports game context without pre-discretization. And possession-based GIM also dominates GIM-T1 after the first season half, which shows the value of including play history in the game context. 
% But learning the history dependence requires a sufficiently large number of observations.
But how quickly and how much the GIM metrics improve depends on the specific success measure. For instance, in Figure~\ref{fig:round-by-round-correlation}, GIM's round-by-round correlation with Goal (top right graph) dominates by round 10, while others require a longer time.

\subsection{Future Seasons: Predicting Players' Salary} \label{sec:salary}

In professional sports, a team will give a comprehensive evaluation to players before deciding their contract. The more value players provide, the larger contract they will get. 
%Thus, players' salary in the subsequent season will be influenced by their performance during current season. 
Accordingly, a good %computational 
performance metric should be positively related to the amount of players' {\em future} contract. The NHL regulates when players can renegotiate their contracts, so we focus on players receiving a new contract following the games in our dataset (2015-2016 season). 
% Mapping performance to salaries is a practically important task because it provides an objective standard to guide players and teams in salary negotiations \cite{idson2000team}.

\begin{table}[htb]
\centering
% \caption{Correlation with Players' Contract}
% \label{table:correlation-player-contract}
\resizebox{0.8\columnwidth}{!}{
\begin{tabular}{ccc}
\hline
methods & 2016 to 2017 Season & 2017 to 2018 Season \\ \hline
Plus Minus & 0.177 & 0.225 \\
GAR & 0.328 & 0.372 \\
WAR & 0.328 & 0.372 \\\hline
EG & 0.587 & 0.6 \\
SI & 0.609 & 0.668 \\
GIM-T1 & 0.596 & 0.69 \\ 
\textbf{GIM} & \textbf{0.666} & \textbf{0.763}\\\hline
\end{tabular}
}
\caption{Correlation with Players' Contract}
\label{table:correlation-player-contract}
\end{table}

\begin{figure}[htb]
\begin{minipage}{.5\columnwidth}
\centering\
\subfloat{\label{main:e}\includegraphics[scale=.22]{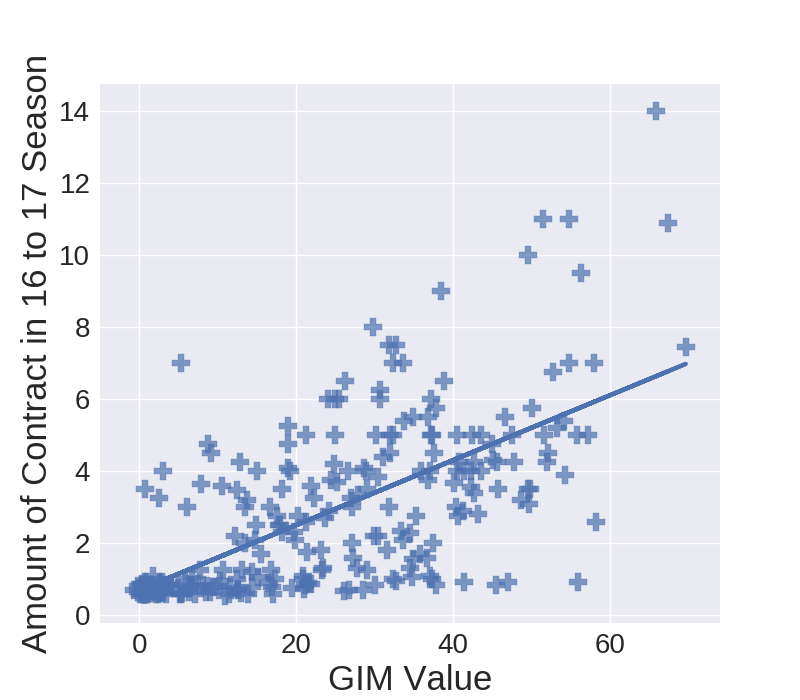}}
\end{minipage}%
\begin{minipage}{.5\columnwidth}
\centering
\subfloat{\label{main:f}\includegraphics[scale=.22]{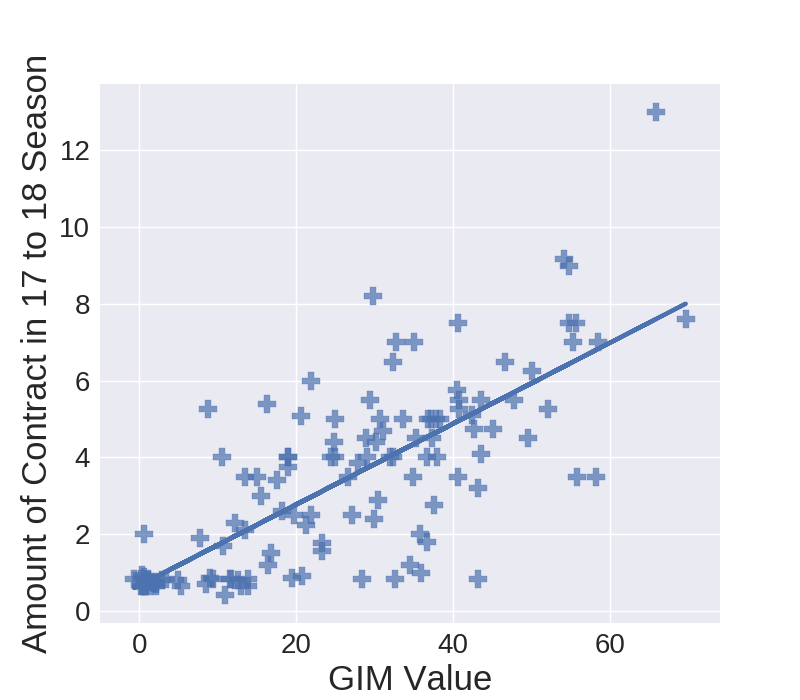}}
\end{minipage}
\caption{Player GIM vs. Value of new contracts in the 2016-17 (left) and 2017-18 (right) NHL season.
\label{fig:scatter-player-contract}}
\end{figure}

Table~\ref{table:correlation-player-contract} shows the metrics' correlations with the amount of players' contract over all the players who obtained a new contract during the 2016-17 and 2017-18 NHL seasons. Our GIM score achieves the highest correlation in both seasons. 
% \textcolor{blue}{
% It provides a better approximation to teams' evaluation over players and thus players' future salary. As GIM also has strong correlation with other succeed metrics and is context-aware and consistent over season, we can use it to quantify player's overall contribution and predict an objective future contract for players.}
%Given the previous validation of GIM, this
This means that the metric can serve as an objective basis for contract negotiations. 
The scatter plots of Figure~\ref{fig:scatter-player-contract} illustrate GIM's correlation with amount of players' future contract. In the 2016-17 season (left), we find many underestimated players in the right bottom part, with high GIM but low salary  in their new contract. 
It is interesting that the percentage of players who are undervalued in their new contract decreases in the next season (from $32/258$ in 2016-17 season to $8/125$ in 2017-2018 season). This suggests that GIM provides an early signal of a player's value after one season, while it often takes teams an additional season to recognize performance enough to award a higher salary. 
%As a informative metric that can model human's subjective evaluation over players, GIM can help to decide the amount of contract a player should receive.

\section{Conclusion and Future Work}
We investigated Deep Reinforcement Learning (DRL) for professional sports analytics. We applied DRL to learn complex spatio-temporal NHL dynamics. The trained neural network  provides a rich source of knowledge about how a team's chance of scoring the next goal %(Q-value) 
depends on the match context. Based on the learned action values, we developed an innovative context-aware performance metric GIM that provides a comprehensive evaluation of NHL players, taking into account {\em all} of their actions. In our experiments, GIM had the highest correlation with most standard success measures, was the most temporally consistent metric, and generalized best to players' future salary. 
% And generalizing from a sample of season matches, GIM was the best predictor of season total success measures. 
Our approach applies to similar continuous-flow sports games with rich game contexts, like soccer and basketball. A limitation of our approach is that players get credit only for recorded individual actions. An influential approach to extend credit to all players on the rink has been based on regression~\cite{Macdonald2011,Thomas2013}. A promising direction for future work is to combine Q-values with regression.\\
% , which enjoy popularity and a large market.
%Similar to ice hockey, many team sports like soccer, basketball also has high popularity, large market and more importantly, complex game context. 
%For the future work, in guiding teams to draft, sign or trade players, we will build 
%A direction for future work is building a DRL model with tracking data to compute GIM scores for players in these sports. 

\section*{Acknowledgements}
This work was supported by an Engage Grant from the National Sciences and Engineering Council of Canada, and a GPU donation from NVIDIA Corporation.

\bibliographystyle{named}
\bibliography{ijcai18}

\end{document}